\documentclass{article}
\usepackage{spconf,amsmath,graphicx}

\usepackage{balance,subcaption}

\usepackage{url}

\usepackage{color}

\title{Visual Features for Context-Aware Speech Recognition}
\name{Abhinav Gupta, Yajie Miao, Leonardo Neves, and Florian Metze\thanks{\hspace{-5.85mm} This  work   used  the  Extreme  Science   and  Engineering  Discovery
Environment (XSEDE), which is supported by National Science Foundation
grant number OCI-1053575. \newline Submitted for publication in fond memory of Yajie Miao ($\dagger$ 2016).}}
\address{Language Technologies Institute, Carnegie Mellon University\\
  Pittsburgh, PA; U.S.A.\\
  \texttt{abhinavgupta94@gmail.com, \{ymiao|lneves|fmetze\}@cs.cmu.edu}}
\begin{document}
%\ninept
%
\maketitle
\begin{abstract}

  \noindent Automatic transcriptions of consumer generated multi-media content such
  as ``Youtube'' videos still exhibit high word error rates. Such data typically occupies a very
  broad domain, has been recorded in challenging conditions, with
  cheap hardware and a focus on the visual modality, and may have been
  post-processed or edited.

  In this paper, we extend our earlier work on adapting the acoustic model
  of a DNN-based speech recognition system to an RNN language model, and
  show how both can be adapted to the objects and scenes that can be
  automatically detected in the video. We are working on a corpus of
  ``how-to'' videos from the web, and the idea is that an object that
  can be seen (``car''), or a scene that is being detected (``kitchen'')
  can be used to condition both models on the ``context'' of the recording,
  thereby reducing perplexity and improving transcription. We achieve
  good improvements in both cases, and compare and analyze the respective
  reductions in word error rate.

  We expect that our results
  can be useful for any type of speech processing in which ``context'' information is available,
  for example in robotics, man machine interaction, or when indexing large audio-visual archives,
  and should ultimately help to bring together the ``video-to-text'' and ``speech-to-text'' communities. 

\end{abstract}
%
%\vspace{-2mm}
\section{Introduction}
\label{sec:intro}
%\vspace{-2mm}

Robustness or adaptation to signal variability is a key challenge if
automatic speech recognition (ASR) systems are to become universally useful. One way
in which this could be achieved is to adapt both the acoustic model and the language
model to the broad ``context'' of the input. By ``context'', we mean essentially
anything that is known about the input speech.

State-of-the-art recognition accuracy on a wide range of acoustic modeling tasks
is defined by DNNs \cite{1,2,3}, or variants thereof. On consumer-generated
content (``Youtube videos'') however, even DNN models exhibit word error rates (WERs)
above 40\% \cite{liao2013large}, although no standardized test set exists.
Other work reports significantly lower \cite{miao:is2016} or higher WERs \cite{florian:icme2014},
showing the wide variability that exists in such data. Most recently, low word
error rates have been recorded in an extremely high resource setup~\cite{45675}.

%DNNs display generally
%better recognition accuracy than traditional Gaussian mixture models (GMMs) \cite{4}.
%Robustness however remains a challenge for DNN models \cite{5}. \cite{6} for
%example shows that the performance of simple DNNs degrades significantly as the signal-to-noise
%ratio (SNR) drops.

An effective strategy to deal with variability is to incorporate
additional, longer-term knowledge explicitly into DNN models: \cite{7,8,9,dnnsat:is2014,yajie:taslp2015}
study the incorporation of speaker-level i-vectors to smooth out the
effect of speaker variability. Time Delay Neural Networks \cite{tdnn,vijay} use wide
temporal input windows to improve robustness dynamically. \cite{vesel2016sequence} extracts
long-term averages from the audio signal to adapt a DNN acoustic model. Similarly,
in \cite{yajie-robust:is2015}, we learn a DNN-based extractor to model the
speaker-microphone distance information dynamically on the frame level. Then
distance-aware DNNs are built by appending these descriptors to the DNN inputs.

It is an important distinction that our work \textsl{does not require} localization of lip regions
and/ or extraction of frame-synchronous visual features (lip contours, mouth shape, SIFT, landmarks, etc.),
as is the case in ``traditional''
audio-visual ASR \cite{13,14,17,20,19}, which has been developed mostly with a focus
on noise robustness. For the majority of our data, lip-related information is not available at all,
or the quality is extremely poor.

In this paper, we present an extension of and comparison to previously published work~\cite{miao:is2016},
by adapting not only the acoustic model, but also the language model of an ASR system to the visual ``context''
that is present in the video stream of open-domain Internet video.
%aim to relax these constraints and demonstrate the effectiveness of adapting both
%the acoustic and the language model of a multi-modal ASR system by using ``context'' information from
%challenging, open-domain Internet videos.
Our approach is based on deep learning, and involves two major steps:

First, we extract visual features using deep Convolutional Neural Networks (CNNs) trained for
object recognition and scene labeling tasks. We extract such information from a single, random frame within
an utterance only, rather than at the level of each frame, but other levels of granularity, or
smoothing approaches are also possible.
Thus, we do not require perfect alignment between audio and video channels, which is often
almost impossible to achieve on data that has been collected ``in the wild''. Our ``context vector''
is therefore an n-dimensional representation of the visuals which are present while an utterance is being spoken.

Then, we adapt the acoustic model of the recognizer using a framework in which the residual error
at the feature inputs of a DNN is reduced with an adaptation network. This network is trained on the context
vector, and predicts a linear shift of the main DNN's input features. The central idea is somewhat
similar to ResNets \cite{resnet}, and was originally developed
for i-vector based adaptation \cite{dnnsat:is2014}. It works well for adaptation to other knowledge sources as well.

For first-pass decoding, we use an in-domain 3-gram language model. To adapt the language model of the recognizer,
we then re-rank 30-best lists with an RNN language model, which has been conditioned on the same segment-level
``context vector'' as the acoustic model. We show that this approach results in significant reductions in
perplexity and also reduces word error rate.

%Audio-visual speech recognition, \cite{miao:is2016,eesen,yajie:taslp2015,yajie-robust:is2015}. \cite{a}

\section{Extraction of Visual Features}

The extraction of visual features follows our previous work on adaptation of DNNs using speaker attributes
\cite{yajie:slt2014} and visual features \cite{miao:is2016}.

Suppose we are dealing with an utterance $u$, which has the acoustic features $\mathbf{O} = \{o_1 , o_2 , ..., o_T \}$, where $T$ is the total number of speech frames. On a video transcribing task, there always exists a video segment corresponding to $u$. This video segment is represented as $\mathbf{V} = \{v_1 , v_2 , ..., v_N \}$, where $N$ is the number of video frames. The video frames are sampled normally at a lower sampling rate than the speech frames, i.e. $N < T$. From this segment, we randomly select a frame $v_n$ which serves as the image representation for the utterance. Then two types of image features are extracted from $v_n$.

\subsection{Object Features}
\label{sec:objectfeatures}

Our first type of visual information is derived from object recognition, a task on which deep learning has accomplished tremendous success \cite{25}. The intuition is that object features contain information regarding the acoustic environment. For example, classifying an image to the classes ``computer keyboard'' and ``monitor'' indicates that the speech segment has been recorded in an office.

We extract this object information using a deep CNN model which has been trained on a comprehensive object recognition dataset, a 1.2 million image subset of ImageNet \cite{26} used for the 2012 ILSVR challenge, and the resulting CNN model is referred to as \textsc{Object-CNN}. Then, on our target ASR task, the video frame $v_n$ is fed into the CNN model, from which we get the distribution (posterior probabilities) over the object classes. These probabilities encode the object-related information that are finally incorporated into DNN acoustic models.
%Due to the high dimension of the probabilities, we may need to perform dimension reduction. More details regarding training of the CNN networks will be presented in Section~\ref{4.1}.

The \textsc{Object-CNN} network follows the standard AlexNet architecture [25]. The network contains 5 convolution layers which use the rectifier non-linearity (ReLU) \cite{28} as the activation function. In the first and second convolution layers, a local response normalization (LRN) layer is added after the ReLU activation, and a max pooling layer follows the LRN layer. In the third and fourth convolution layers, we do not apply the LRN and pooling layers. In the fifth convolution layer, we only apply the max pooling layer, without LRN being applied. 3 fully-connected (FC) layers are placed on top of the convolution layers. The first and second FC layers have 4096 neurons, whereas the number of neurons in the last FC layer is equal to the number of classes, 1000 in our case.
%The \textsc{Object-CNN} is trained on ImageNet [26], a dataset containing over 15 million labeled images belonging to around 22,000 categories. We use a subset of ImageNet created for the 2012 Large-Scale Visual Recognition Challenge (ILSVRC). The ILSVRC training set amounts to 1.2 million images cover- ing 1000 classes. Each image from the training data is resized to 256x256, and then normalized with mean and variance normal- ization. From this resized image, we crop the 227x227 block from the center as the inputs into the CNN. Therefore, the CNN inputs have the size of 3x227x227, where 3 is the number of channels (RGB).
Model training optimizes the standard cross-entropy (CE) objective. The resulting \textsc{Object-CNN} achieves a 20\% top-5 error rate on the ILSVRC 2012 testing set.

\subsection{Place Features}

The utility of the object features comes from the ``place'' information that is implicitly encoded by the object classification results. It is then natural to utilize place features in a more explicit way. To achieve this, we train a deep CNN model meant for the scene labeling task. Given a video frame, the classification outputs from this \textsc{Place-CNN} encode the place information, which is then incorporated into acoustic models. For convenience of formulation, the resulting visual feature vector for this utterance $u$ is represented as $f_u$.

In order to extract place information, we train the \textsc{Place-CNN} network on the MIT Places dataset \cite{29} which contains 2.5 million images belonging to 205 scene categories. Examples of the scenes include ``dining room'', ``coast'', ``conference center'', ``courtyard'', etc. We use the complete set of 2.5 million images for training, and follow the same image pre-processing as used on ImageNet (Section~\ref{sec:objectfeatures}). The architecture of the \textsc{Place-CNN} is almost the same as that of the \textsc{Object-CNN}. The only difference is that in the final FC layer, the \textsc{Place-CNN} has 205 neurons corresponding to the 205 scene classes, whereas the \textsc{Object-CNN} contains 1000 neurons.

\section{Experimental Setup}
\label{data}

We chose to investigate context-aware ASR on a dataset of real-world English instructional videos, which we had downloaded from online video archives \cite{yajie-robust:is2015,yajie:slt2014}. These videos have been uploaded by social media users to share expertise on specific tasks (e.g., oil change, sandwich making, etc.). ASR on these videos is challenging because they have been recorded in various environments (e.g., office, kitchen, baseball field, train, etc.), giving us a variety of contexts, yet they are rich in speech, making them suitable for the proposed work. Our main training set comprises 90 hours of speech (3900 videos), and we use 4 hours for testing (156 videos). 

We used Kaldi~\cite{kaldi} and PDNN~\cite{pdnn} for our experiments, training a 5-layer DNN acoustic model using cross-entropy \cite{miao:is2016}. For decoding, a trigram language model (LM) is trained on the training transcripts. This LM is then interpolated with another trigram LM trained on an additional set of 270 hours transcriptions of instructional videos. The complete set of 360 hours is also used for training the RNN language model.

\section{Acoustic Model Adaptation}
\label{amadapt}

In previous work \cite{dnnsat:is2014,yajie:taslp2015}, we presented a framework to perform speaker adaptive training (SAT) for DNN models. This approach requires an i-vector \cite{27} to be extracted for each speaker. Based on the well-trained speaker-independent (SI) DNN, a separate adaptation neural network is learned to convert i-vectors into speaker-specific linear feature shifts. Adding these shifts to the original DNN inputs produces a speaker-normalized feature space. Parameters of the SI-DNN are re-updated in this new space, generating the SAT-DNN model. This framework has also been applied successfully to descriptors of speaker-microphone distance \cite{yajie-robust:is2015}, and we find it to be more robust than straightforward feature concatenation \cite{yajie:taslp2015}.

We port this idea to visual input features, which enables us to conduct ``context'' adaptation for DNNs, simply replacing the i-vector representation with the visual features. An adaptation network is learned to take the visual features as inputs and generate an adaptive feature space with respect to the visual descriptors. Note that in this case, the linear feature shifts generated by the adaptation network are utterance-specific rather than speaker-specific. Re-updating the parameters of the DNN in the normalized feature space gives us the adaptively trained ``video adaptive training'' VAT-DNN model \cite{miao:is2016}. This VAT-DNN model takes advantage of the visual features as additional knowledge, and generalizes better to unseen variability. In our setup, we generate 100-dimensional utterance-level visual ``context'' features by projecting the output vector (after soft-max) of the \textsc{Place-CNN} and \textsc{Object-CNN} (either individually, or in concatenation) down to 100 dimensions using Principal Component Analysis (PCA).
%(100-dimensional PCA projections of activations in the FC5 and/ or FC7 layers) are taken as the inputs into the adaptation network, which contains 3 hidden layers with 512 neurons per layer.
The outputs of the adaptation network are 40-dimensional shifts to lMEL features.

\section{Language Model Adaptation}
\label{lmadapt}

To adapt the language model, we used the same features that we also use for adapting the
acoustic model (100-dimensional PCA projections of place and scene) as ``topic'' information
in a context dependent Recurrent Neural Network (RNN)
language model~\cite{mikolov2012context}. The vocabulary contains about 35k words.
A two-layer bidirectional
LSTM with an embedding layer size of 900 and 1024 cells per layer gave lowest perpexities
in initial cross-validation experiments
on 360 hours of data. This architecture was thus adopted for the
experiments below. We used tanh non-linearities and 0.5 as the dropout factor, without any
additional regularization
except gradient clipping (at 100). The initial learning rate was 0.01; training used
Adagrad \cite{adagrad} and a batch size of 128. The network was implemented in Lasagne \cite{lasagne}.

We provide the adaptation vector only at the beginning of the sentence, although it might make
sense to provide it also at intermediate steps, as the average sentence length is 18 words.

\section{Experiments}
\label{sec:exps}

To reduce the dimensionality of the adaptation feature and to facilitate comparison with earlier work on i-vector adaptation, we reduce the dimensionality of the place and object features to 100 (from 1000 and 205) using PCA, estimated on the training part only of the audio-visual dataset.

\begin{table}
  \begin{tabular}{ccccccc}
    \hline
    Base-  & Object   & Place    & O. + P. & i-      & All \\
    line   & F.       & F.       & F.      & vectors & Features \\
    \hline
    23.4\% & 22.5\% & 22.5\% & 22.3\% & 22.0\% & 21.5\%  \\
    \hline
  \end{tabular}
  \caption{Word error rates when applying acoustic model adaptation \label{tab:am} using
    object features, place features, a combination thereof
    ~\cite{miao:is2016}, i-vectors~\cite{miao:is2016}, and a combination of the
  two visual features with i-vectors (``all'').}
\end{table}

Table~\ref{tab:am} shows the result of adapting the DNN acoustic model with visual features,
and i-vectors for comparison, as well as a combination of visual features and i-vectors. Gains
are consistent, and quite complementary when using the concatenation of visual features and i-vectors
for adaptation. Also, in all cases, the adaptation network method outperforms simply concatenating
the adaptation vector to the input features.

\begin{figure}
  \centering
  \includegraphics[width=\columnwidth,clip=true,trim=5cm 10.5cm 4.5cm 10.7cm]{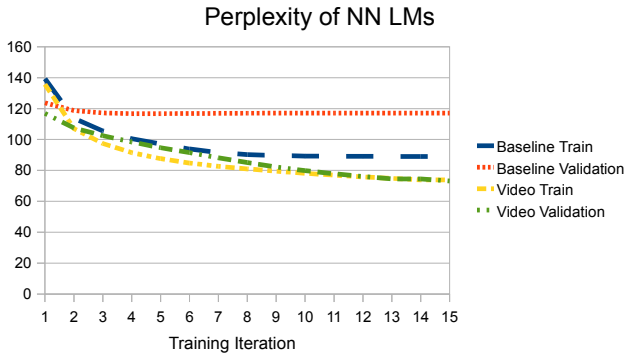}
  % trim = left bottom right top
  \caption{Training and validation perplexity of the NN LMs trained with and without
    visual features, averaged across 5 folds, on 360 hours of data. Parameters were
    optimized for the adaptive LMs, hence the baseline LM converges quickly. Validation
    perplexity is lower than training perplexity initially because it is measured at the end
    of each training iteration, while training perplexity is being computed while processing
    the data and updating the model. \label{fig:nnlm}}
\end{figure}

Next, we use the same method to adapt the language model to the visual information. To find the
best meta parameters for the LSTM language model, we performed 5-fold cross-validation on the entire
360 hours of training data, and averaged the results. Figure~\ref{fig:nnlm} shows that conditioning
the LSTM LM on video features reduces perplexity from 89 to 74 on training data, which is a significant
reduction, which we find does also carry over to the word error rates on unseen test data. 

We generated 30-best lists using the baseline acoustic models (with a WER of 23.4\%),
which had an oracle WER of 15.6\%, and re-ranked them with all 5 neural network language models (NN LMs),
averaging the language model scores. Using the concatenation of object and place features as inputs to the NN LM,
we achieve a word error rate of 22.6\%, which is very close to the performance achieved with
the adaptation of the acoustic model.

\section{Analysis of Results}
\label{sec:analysis}
% This is in /data/ASR5/fmetze/Abhinav-Youtube-Exps and and /home/abhinav5 on rocks

For both acoustic and language model adaptation, we performed some more in-depth analysis
to see where gains are mostly coming from.

\begin{figure}
  \centering
  \begin{subfigure}{.5\linewidth}
    \centering
    \includegraphics[width=0.95\linewidth,clip,trim=0 9mm 0 9mm]{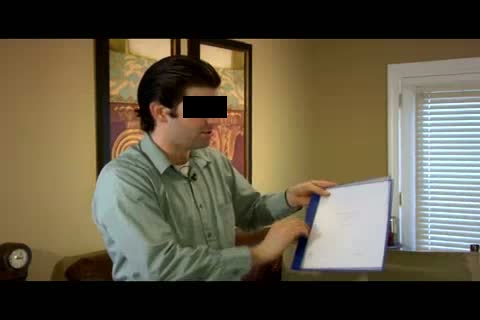}
%    \caption{A subfigure}
%    \label{fig:sub1}
  \end{subfigure}%
  \begin{subfigure}{.5\linewidth}
    \centering
    \includegraphics[width=0.95\linewidth]{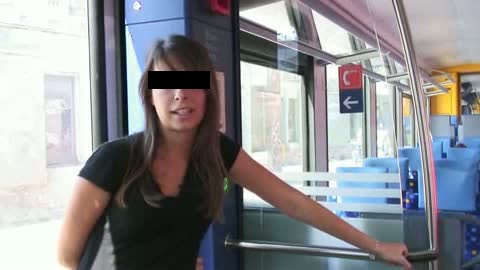}
%    \caption{A subfigure}
%    \label{fig:sub2}
  \end{subfigure}
  \caption{Keyframes for two typical videos in our dataset. The baseline WER is 27.6\% for the
    ``home'' video on the left , and 47.7\% for the ``train'' video on the right. Acoustic model
    adaptation does not improve the ``home'' video at all, but reduces WER to 38.2\% on the
    ``train'' video. Language model adaptation
    improves WER on both videos slightly, to 26.6\% and 43.2\% respectively. \label{fig:images}}
\end{figure}

We manually inspected those videos on which we observed more than 10\% relative WER reductions,
and find that they have been recorded either in outdoor environments (e.g.\,baseball field,
airport apron, street, etc.), or in non-typical indoor conditions (e.g.\,kitchen, music studio, etc.),
where music/ noise may interfere with the actual speech a lot. Adding the scene descriptors helps the
DNN model normalize the acoustic characteristics of these rare conditions, and thus benefits the
generalization to unseen testing speech. We then labeled all 156 testing videos as either
``typical indoors'' (e.g.\,office) and ``other'' (noisy indoors, outdoors), and analyzed the
relative improvements with the of a system adapted with \textsc{Place-CNN} features only,
and find that the ``quiet indoors'' videos get improved from 22.1\% WER to 21.7\%, while ``other''
videos get improved from 27.6\% to 25.7\%. ``Other'' videos thus get improved by 7\% relative, while
clean videos get improved by 2\% only.

When training the NN LM on 90 hours of data only, adaptation with \textsc{Object-CNN} features
results in a perplexity of 94.7, while adaptation with \textsc{Place-CNN} features gives a
perplexity of 98.9. It seems intuitive that ``objects'' would be slightly more salient for the
topic of a ``how-to'' video than the scene.
% object-place.ods

% TRAIN video with picture is G4WcRgmOXI4(?)
% baseline -> am-adapt (from thesis) -> lm-adapt (from here)
% 27.55 -> 27.55 -> 26.63
% 47.71 -> 38.24 -> 43.24

Figure~\ref{fig:images} shows typical keyframes from our database, and the typical pattern of
improvements: acoustic model adaptation tends to give significant improvements on ``outdoor''
videos only (55 videos), while language model adaptation tends to give smaller improvements across
the board (only 30 of 156 videos deteriorate).

%\vspace{-2mm}
\section{Conclusion and Future Work}
%\vspace{-2mm}

In this paper, we described a system that extracts context information that is relevant
for speech processing from the visual channel of the video. We showed that the information
can be incorporated in both acoustic \textsl{and} language models, and that this approach
leads to systematic and consistent improvements. These observations are in line with recent work
on multi-modal machine translation \cite{2016arXiv160500459E}.

We are currently expanding the acoustic model adaptation experiments to the larger (360 hours)
version of the corpus, and expect to see further performance improvements by combining both acoustic
and language model adaptation.
We are also experimenting with different and better ways of incorporating the video features into the
language model, and attempt more insightful analysis of the results, e.g. how much do different
types of features contribute to the different models (e.g., do the scene features contribute
relatively more to the acoustics, while the object features contribute more to the language model?),
and what types of errors are being reduced (e.g., nouns? verbs? semantically confusing errors?).

In the long term, this work should help to improve fully end-to-end ``video-to-text'' approaches,
which generate image or video ``summaries'' based on multi-modal embeddings, and reference ``captions''
\cite{kiros14,venugopalan2014translating,vinyals2015show}, rather than speech recognition transcriptions.


%\clearpage
\balance
\ninept

\bibliographystyle{IEEEbib}
\bibliography{strings,refs,../bibtex/florian}

\end{document}